\documentclass[11pt]{article}

\usepackage{epsfig,amsmath,latexsym,amssymb}
\usepackage{graphicx}
\usepackage{lscape}
\usepackage{picture, eso-pic, tikz} %% for gray boxes
\usepackage{dsfont}
\usepackage{listings}
\usepackage{changes}
\usepackage{hyperref}
\usepackage{bbding}

\renewcommand{\baselinestretch}{1.1}
\def\R{{\mathbb R}}  %%
\def\p{{\mathbb P}}  %% 
\def\E{{\mathbb E}}  %
\def\Beweis{\footnotesize}

\newcommand{\Remm}[1]{}
\newtheorem{theo}{Theorem}[section]

\newtheorem{prop}[theo]{Proposition}

\newtheorem{model ass}[theo]{Model Assumptions}

\newtheorem{example}[theo]{Example}

\newtheorem{rem}[theo]{Remark}
\def\EndProof{\hfill {\scriptsize $\Box$}}
\def\EndExample{\hfill {\scriptsize $\blacksquare$}}
\numberwithin{equation}{section}

\definecolor{MyGray}{rgb}{0.92,0.92,0.92}


\newcommand{\bl}[1]{\textcolor{blue}{{#1}}}

\definecolor{British racing}{rgb}{0.0, 0.5, 0.0}

\def\bX{\boldsymbol{X}}
\def\bV{\boldsymbol{V}}
\def\b0{\boldsymbol{0}}

\def\bd{\boldsymbol{d}}

\def\b0{\boldsymbol{0}}

\def\bD{\boldsymbol{D}}
\def\bU{\boldsymbol{U}}

\usepackage{todonotes}
\newcommand{\Comments}{1}
\newcommand{\mynote}[2]{\ifnum\Comments=1\textcolor{#1}{#2}\fi}
\newcommand{\mytodo}[2]{\ifnum\Comments=1%
  \todo[linecolor=#1!80!black,backgroundcolor=#1,bordercolor=#1!80!black]{#2}\fi}

\ifnum\Comments=1               % fix margins for todonotes
\fi

\ifnum\Comments=1               % fix margins for todonotes
\fi

\ifnum\Comments=1               % fix margins for todonotes
\fi

\begin{document}
\author{Mathias Lindholm
\quad Ronald Richman
\quad Andreas Tsanakas
\quad Mario V.~W\"uthrich}

\date{Version of \today}
\title{{\sc A Discussion of Discrimination and Fairness in Insurance Pricing}}
\maketitle

\begin{abstract}
\noindent  
Indirect discrimination is an issue of major concern in algorithmic models.
This is particularly the
  case in insurance pricing where protected policyholder characteristics are
  not allowed to be used for insurance pricing. Simply disregarding protected policyholder information
  is not an appropriate solution because this
  still allows for the possibility of inferring the
  protected characteristics from the non-protected ones. This leads to
  so-called proxy or indirect discrimination. Though proxy discrimination 
  is qualitatively different from the group fairness concepts in machine learning, these
  group fairness concepts are proposed to `smooth out' 
  the impact of protected characteristics in the calculation of insurance prices.
  The purpose of this note is to share some thoughts about group fairness concepts
  in the light of insurance pricing and to discuss their implications. We present a statistical model that is 
  free of proxy discrimination, thus, unproblematic from an insurance pricing point of view.
  However, we find that the canonical price in this statistical model does not
  satisfy any of the three most popular group fairness axioms. This seems puzzling and
  we welcome feedback on our example and on the usefulness of these
  group fairness axioms for non-discriminatory insurance pricing.

~


\end{abstract}

\section{Introduction}
For legal and for societal reasons, there are several policyholder attributes that
are not allowed to be used for insurance pricing, e.g., the European Council \cite{EuropeanCouncil,
  Europe1} does not allow for the use of gender information in insurance pricing, or ethnicity
is a critical attribute that may be declared as protected information by society.
Frees--Huang \cite{FreesHuang} and Xin--Huang \cite{XinHuang} give extensive overviews on
protected information in insurance and implications for pricing, while
Avraham et al.~\cite{Avraham} and Prince--Schwarcz \cite{PrinceSchwarcz} provide 
legal viewpoints on this topic. The critical issue is that just ignoring protected
information (being unaware of protected information) does not solve the problem, as
protected information can be inferred from non-protected one if corresponding
variables are correlated. This is especially true
in high-dimensional algorithmic models. Such inference is called proxy discrimination
or indirect discrimination, and it is often implicitly performed  
during the fitting procedure of complex models.

There are several attempts to prevent 
this inference, e.g.,
there is a counterfactual approach from causal statistics,
see Kusner et al.~\cite{Kusner} and Araiza Iturria et al.~\cite{AraizaIturria},
there is a probabilistic approach called discrimination-free
insurance pricing by Lindholm et al.~\cite{Lindholm1, Lindholm2},
and there are group fairness approaches inspired by
machine learning concepts, see Grari et al.~\cite{Grari}.
The purpose of this note is to discuss the three most popular group fairness axioms
in the light of insurance pricing. We define a statistical model
that is free of discrimination because the non-protected information fully describes
the distribution of the response variable, and the protected one does not carry any additional information
about the response.
We analyze the three
group fairness axioms in this model. Surprisingly, the canonical
non-discriminatory price (which is uncontroversial here) does not satisfy any of these three group fairness
axioms. This seems puzzling and it requires a broader discussion within
our community about discrimination and fairness in insurance pricing. Therefore, we share our thoughts
and welcome feedback on our example and on the use
of the group fairness axioms for non-discriminatory pricing.

\bigskip

{\bf Organization.}
We introduce our statistical model that is free of discrimination
in Section \ref{Individual fairness}, and we calculate
the best-estimate and the unawareness prices in this model.
In Section \ref{Group fairness}, we formally introduce the
three most popular group fairness axioms; the independence axiom (statistical parity),
the separation axiom (equalized odds) and the sufficiency axiom (predictive parity); see
Barocas et al.~\cite{Barocas}, Xin--Huang \cite{XinHuang}
and Grari et al.~\cite{Grari}. We show that in our model the canonical
non-discriminatory price does not satisfy any of these three axioms.
Finally, in Section \ref{Conclusions and open questions} we conclude
and discuss further issues. All mathematical results are proved
in Appendix \ref{Appendix: mathematical proofs}.

\section{Running example}
\label{Individual fairness}
To set the ground, we fix a probability space $(\Omega, {\cal F}, \p)$ with $\p$ describing the
real world probability measure. On this probability space we consider the random vector
$(Y,\bX,\bD)$. The response $Y$ describes the insurance claim that we try to predict (and price).
The vector $\bX$ describes the {\it non-protected characteristics} ({\it non-discriminatory covariates}),
and $\bD$ describes the {\it protected characteristics} ({\it discriminatory
covariates}). We assume that the partition
of the covariates into protected and non-protected ones is given exogenously, e.g., by law
or by societal norms and preferences. We use the distribution $\p(\bX, \bD)$ 
to describe the insurance portfolio, i.e., the random selection
of a policy from the insurance portfolio. Different insurance companies may have 
different such distributions, and the insurance portfolio distribution $\p(\bX, \bD)$
typically differs from the overall population distribution in a given society.

\paragraph{Best-estimate price.}
For insurance pricing, one designs a regression model
that describes the conditional distribution of $Y$, given 
the explanatory variables  $(\bX, \bD)$. 
The {\it best-estimate price} of $Y$, given full information $(\bX, \bD)$, is given by
\begin{equation}\label{def: best-estimate price}
\mu(\bX, \bD) = \E \left[\left. Y\right|\bX, \bD \right].
\end{equation}
This price is called best-estimate because it has minimal prediction variance,\footnote{For simplicity, we
assume that all considered random variables are square-integrable w.r.t.~$\p$.}
i.e., it
is the most accurate predictor for $Y$, given $(\bX,\bD)$, in the $L^2$-sense.

\medskip

\bl{\HandRight} ~ In general, the best-estimate price {\it directly discriminates} because it uses the protected information $\bD$
as an input.

\paragraph{Fairness through unawareness.} The most simple fairness concept in machine learning
is the {\it fairness through unawareness} concept that just drops the protected information $\bD$.
This results in the {\it unawareness price} of $Y$, given $\bX$, defined by
\begin{equation}\label{def: unawareness price}
\mu(\bX) = \E \left[\left. Y\right|\bX \right].
\end{equation}
The unawareness price does not {\it directly discriminate} because it does not use $\bD$ as 
an input, i.e., it is blind w.r.t.~$\bD$. 
However, it may {\it indirectly discriminate} because the knowledge of $\bX$
allows inference of $\bD$ through the tower property of conditional
expectation
\begin{equation}\label{formula indirect discrimination}
\mu(\bX) = \int \mu(\bX, \bd) \, d\p(\bD=\bd |\bX). 
\end{equation}
This formula shows that if there is statistical dependence between $\bD$ and $\bX$ w.r.t.~$\p$,
we implicitly use this dependence for inference, see  last term in \eqref{formula indirect discrimination}.
This is what is called  proxy discrimination and indirect discrimination, and it should
be avoided.

\medskip

\bl{\HandRight} ~ Fairness through unawareness indirectly  discriminates, and it does not solve the problem of non-discriminatory
insurance pricing.

\medskip

We now present our example. In this example we have a response variable $Y$ whose
distribution function is fully described by the non-protected information $\bX$, and
the protected information $\bD$ does not carry any additional information about the
response $Y$.\footnote{\label{not a proper risk factor}Araiza Iturria et al.~\cite{AraizaIturria} call such a variable $\bD$ not a true risk factor,
because $\bX$ is sufficient to describe the distribution of $Y$, see also European Commission \cite{Europe1}.}
Therefore, we consider this statistical model to be
free of discrimination.
This model is simple enough to be able to calculate all terms
of interest, and, even if it is unrealistic in practice, it allows us to draw conclusions
about the group fairness axioms and discrimination in insurance pricing.

\begin{example}[running example]\normalfont
  \label{Example}
Assume we have three-dimensional covariates
$(\bX, \bD) = (X_1,X_2,D)$ having a multi-variate Gaussian
portfolio distribution
\begin{equation}\label{Gaussian model}
(X_1,X_2,D)^\top ~\sim~{\cal N} 
\left(\begin{pmatrix}0\\0\\0\end{pmatrix}, \,\Sigma=
\begin{pmatrix}
1&0&\rho_1\\
0&1&\rho_2\\
\rho_1&\rho_2&1
\end{pmatrix}
=\begin{pmatrix}
1&0&0.1\\
0&1&0.9\\
0.1&0.9&1
\end{pmatrix}
\right).
\end{equation}
Thus, the non-protected covariates $X_1$ and $X_2$ are independent,
but they are both dependent with the protected covariate $D$. We assume that
this dependence is a purely statistical one implied by the portfolio
structure of the considered insurance company, and not a causal one. That is,
there might be a second insurance company with a portfolio distribution $\p(\bX,\bD)$
described by a Gaussian distribution \eqref{Gaussian model}, but having, e.g.,  negative correlations
between $\bX$ and $\bD$.

For the
response $Y$ we assume that conditionally, given $(\bX,\bD)$,
\begin{equation}\label{Gaussian sufficiency}
Y|_{(\bX,\bD)} ~\sim~{\cal N}(X_1, 1+X_2^2).
\end{equation}
That is, the response does {\it not} depend on the protected information $\bD$, but only
on the non-protected information $\bX$. This means that $\bX$ is sufficient to describe
the distribution of $Y$, and $\bD$ does not carry any additional information about $Y$,
see also Footnote \ref{not a proper risk factor}.

The best-estimate and the unawareness prices coincide in this example, and they are given by
\begin{equation}\label{these prices are not sufficient}
\mu(\bX,\bD) =\mu(\bX)= X_1.
\end{equation}
In this example, there is no indirect discrimination through a conditional
expectation \eqref{formula indirect discrimination} because the best-estimate price
coincides with the unawareness price, and the explicit structure of the portfolio
distribution $\p(\bX,\bD)$ is {\it not} needed here.
Therefore, in this
example, we call these prices {\it non-discriminatory}.\footnote{\label{footnote1}More generally, the definition
  of the non-discriminatory price is uncontroversial in insurance in examples where the non-protected information $\bX$ is sufficient
  to describe the response $Y$, and the additional knowledge of the protected information $\bD$ is not needed because it 
  does not reveal any additional information about the response $Y$; we come back to this in
  formula \eqref{remark independence model 0} and in Remark \ref{global sufficiency}, below.}
From an actuarial viewpoint this price is non-discriminatory because the protected
information $\bD$ has no influence on the response $Y$, given $\bX$.
This is true for {\it any} dependence structure between $\bD$ and $\bX$,
and no matter whether it is a causal or a purely statistical one.
This is an additional justification for not having indirect discrimination.
\EndExample
\end{example}

\begin{rem}\normalfont
  Under the assumptions of Example \ref{Example}, the non-discriminatory price 
  $\mu(\bX)$ given in \eqref{these prices are not sufficient} is equal to the discrimination-free insurance price of Lindholm et al.~\cite{Lindholm1}.
  \end{rem}

\section{Group fairness axioms}
\label{Group fairness}
In this section, we introduce the three most popular group fairness axioms
from machine learning; we also refer to
Barocas et al.~\cite{Barocas}, Xin--Huang \cite{XinHuang}
and Grari et al.~\cite{Grari}. We will show 
that the non-discriminatory price of Example \ref{Example} given in
\eqref{these prices are not sufficient} violates all three of these group fairness axioms.

\medskip

We denote by $\widehat{\mu}(\bX)$ any $\sigma(\bX)$-measurable predictor of $Y$, which can
be the unawareness price \eqref{def: unawareness price} or any other pricing functional that solely depends on
the non-protected information $\bX$.

\paragraph{(i) Independence axiom / statistical parity.} 
\label{Statistical parity (independence axiom)} 
Statistical parity is also called demographic parity, and following
Definition 1 of Agarwal et al.~\cite{Agarwal} it is given as follows:
We have {\it statistical parity} if
\begin{center}
$\widehat{\mu}(\bX)$ and $\bD$ are independent under $\p$.
\end{center}
This independence implies for the distribution of the insurance prices $\widehat{\mu}(\bX)$, a.s., 
\begin{equation}\label{statistical parity formula}
\p \left[ \left. \widehat{\mu}(\bX) \le m \right| \bD  \right]
~=~\p \left[  \widehat{\mu}(\bX) \le m  \right]
\qquad \text{for all $m\in \R$.}
\end{equation}

We have the following proposition.

\begin{prop}\label{proposition no independence}
The non-discriminatory price $\mu(\bX)=X_1$ of Example \ref{Example} does not satisfy
the independence axiom (statistical parity).
\end{prop}

\bl{\HandRight} ~ The non-discriminatory price of Example \ref{Example}
violates the independence axiom (statistical parity). This violation is implied by the dependence
of the protected and the non-protected information, which is described
by the parameters $(\rho_1, \rho_2)$ in Example \ref{Example}.
Remark that every insurance company may have a different portfolio distribution
$\p(\bX, \bD)$, e.g., Company 1 may have the portfolio distribution \eqref{Gaussian model} with a
positive correlation $\rho_1=0.1$
and Company 2 may have this portfolio distribution but with independence $\rho_1=0$. The independence axiom 
implies that Company 1 cannot use information $X_1$ for (statistical parity-fair) insurance pricing, whereas Company 2
would be allowed to include information $X_1$ in its tariff. Such a regulation would  be very problematic
as it generates unwanted distortions at the insurance market, in a situation where we have a pricing
problem that is apparently free of discrimination (i.e., in Example \ref{Example}). The only price
that Company 1 could charge under the independence axiom is $\mu=\E[Y]$, which corresponds to the null
model not considering any covariates. However, this is not in line with, e.g., Article 2.3.1(17)
of the guidelines of the European Commission \cite{Europe1} which explicitly allows for the use of covariate $X_1$.
We conclude that the independence axiom (statistical
parity) may be too restrictive in insurance pricing because the use of specific covariates
will differ across insurance companies, and the resulting
(statistical parity-fair) insurance pricing functionals may be impractical.

\begin{rem}\normalfont\label{remark independence model}
    Assume there is a decomposition $\bX=(\bU,\bV)$ such that the protected information $\bD$ is independent 
   from $\bU$
but not from $\bV$. Moreover, as in Example \ref{Example}, assume that, a.s.,
\begin{equation}\label{remark independence model 0}
  \p \left[\left. Y \le y \right| \bX, \bD\right]=\p \left[\left. Y \le y \right| \bX\right] \qquad
  \text{ for all $y\in \R$.}
\end{equation}
In this case, the best-estimate price coincides with the unawareness price, and we
generally set for the non-discriminatory price $\mu(\bX)=\mu(\bX)=\E[Y|\bX]$,
see Footnote \ref{footnote1} on page \pageref{footnote1}.
This non-discriminatory price does not
satisfy the independence axiom in general. On the other hand, the insurance price based only on $\bU$
is equal to $\mu(\bU)=\E[Y|\bU]$, and it  satisfies the independence axiom because $\bU$
and $\bD$ are independent by assumption. If we insist on using the independence axiom,
here, we discard a possibly large part of the non-protected information $\bX$, here $\bV$. This  seems to
be too restrictive, because assumption  \eqref{remark independence model 0} provides a statistical model
that does not require any knowledge about $\bD$ (and of the portfolio distribution $\p(\bX,\bD)$),
and it is also not in line with the guidelines of the European Commission \cite{Europe1}.
\end{rem}

\paragraph{(ii) Separation axiom / equalized odds.} 
Equalized odds is sometimes also called
disparate mistreatment. It has been
introduced by Hardt et al.~\cite{Hardt}, and it is defined as follows:
We have {\it equalized odds} if
\begin{center}
$\widehat{\mu}(\bX)$ and $\bD$ are conditionally independent under $\p$, 
given the  response $Y$.
\end{center}
This conditional independence implies for the distribution of the prices $\widehat{\mu}(\bX)$, a.s, 
\begin{equation}\label{definition of separation axiom}
\p \left[ \left. \widehat{\mu}(\bX) \le m \right| Y, \bD \right]=
\p \left[  \left. \widehat{\mu}(\bX) \le m \right| Y \right]
\qquad \text{for all $m \in \R$.}
\end{equation}

Note that, in general, independence between $\bX$ and $\bD$ is not sufficient to
receive equalized odds.

\begin{prop}\label{proposition no separation}
The non-discriminatory price $\mu(\bX)=X_1$ of Example \ref{Example} does not satisfy
the separation axiom (equalized odds).
\end{prop}

\bl{\HandRight} ~ The non-discriminatory price of Example \ref{Example}
violates the separation axiom (equalized odds). This violation is again implied by the dependence
of the protected and the non-protected information, which is described
by the parameters $(\rho_1, \rho_2)$ in Example \ref{Example}. Therefore, basically, the
same remarks apply as for the failure of the independence axiom.

\paragraph{(iii) Sufficiency axiom / predictive parity.}  
For predictive parity we exchange the role of the response $Y$ and the predictor 
$\widehat{\mu}(\bX)$ compared to equalized odds.
We have {\it predictive parity} if
\begin{center}
$Y$ and $\bD$ are conditionally independent under $\p$, 
given the prediction $\widehat{\mu}(\bX)$.
\end{center}
This conditional independence implies for the distribution of the response $Y$, a.s.,
\begin{equation}\label{sufficient definition}
\p \left[ \left. Y \le y \right| \widehat{\mu}(\bX), \bD  \right]=
\p \left[  \left. Y \le y \right| \widehat{\mu}(\bX) \right]
\qquad \text{for all $y \in \R$.}
\end{equation}

The notion of predictive parity is inspired by the definition of a sufficient statistics in 
statistical estimation theory. We denote the support of the protected information $\bD$ by ${\mathfrak D}$.
We can then interpret 
${\cal P}=\left\{\p_{\bd}[Y\in \,\cdot\,]:=\p[Y \in \,\cdot\,|\bD=\bd];\, \bd \in {\mathfrak D}\right\}$ 
as a family of distributions of $Y$ being parametrized by $\bd \in {\mathfrak D}$. In statistics
we call $\widehat{\mu}(\bX)$ sufficient for ${\cal P}$ if \eqref{sufficient definition} holds.
Basically, this means that $\widehat{\mu}(\bX)$ carries all the necessary information to predict
$Y$, and the explicit knowledge of $\bD=\bd$ is not necessary.

\begin{prop}\label{proposition no sufficiency}
The non-discriminatory price $\mu(\bX)=X_1$ of Example \ref{Example} does not satisfy
the sufficiency axiom (predictive parity).
\end{prop}

\bl{\HandRight} ~ The non-discriminatory price of Example \ref{Example}
violates the sufficiency axiom (predictive parity). This violation is again implied by the dependence
of the protected and the non-protected information, 
and the same remarks apply as for the failure of the independence and the separation axioms.

\begin{rem}\normalfont
  \label{global sufficiency}
    We briefly give the intuition why the sufficiency axiom fails in Example \ref{Example}.
    The sufficiency axiom \eqref{sufficient definition} is formulated w.r.t.~the knowledge of the predictor $\widehat{\mu}(\bX)$. This is different
from a sufficiency condition w.r.t.~the non-protected information $\bX$ where, a.s., 
\begin{equation}\label{sufficient definition for X}
\p \left[ \left. Y \le y \right| \bX, \bD  \right]=
\p \left[  \left. Y \le y \right| \bX \right]
\qquad \text{for all $y \in \R$.}
\end{equation}
The condition \eqref{sufficient definition for X}
has already been considered in \eqref{remark independence model 0}, and
we also refer to Footnote \ref{not a proper risk factor} on page \pageref{not a proper risk factor}.
In general, the information set generated by $\bX$ is bigger than
the information set generated by $\widehat{\mu}(\bX)$. Therefore, the
two sufficiency conditions typically differ. More specifically, 
the price functional $\bX \mapsto \widehat{\mu}(\bX)$
  is a projection of the (multi-dimensional) covariate information $\bX$ to the sub-space
  generated by the one-dimensional prices $\widehat{\mu}(\bX)$. In general, such a projection leads to a loss of information
  which is described by the complement of this projection (in the case of our Example \ref{Example} this is $X_2$).
  If the protected information $\bD$ is dependent with this complement, it will partly compensate for this
  loss of information which results in the failure of the sufficiency axiom (in case this complement
  is needed to fully characterize the distribution of $Y$).
\end{rem}

\section{Discussion and open questions}
\label{Conclusions and open questions}

We have defined a statistical model in Example \ref{Example} where the non-protected information $\bX$ is sufficient
to describe the distribution of the response variable $Y$, and the additional knowledge of the protected information $\bD$ does not
reveal any additional information about this response variable $Y$; Araiza Iturria et al.~\cite{AraizaIturria} call such
a variable $\bD$ not a true risk factor. In this statistical modeling set-up, there is a canonical non-discriminatory
price because the knowledge of $\bD$ is completely irrelevant for best-estimate pricing, in fact, proxy discrimination \eqref{formula indirect discrimination}
does not happen in this set-up.
In Propositions \ref{proposition no independence}, \ref{proposition no separation}
and \ref{proposition no sufficiency} we show that this non-discriminatory price does not satisfy
any of the three group fairness axioms of 'statistical parity', 'equalized odds' and 'predictive parity'.
In other words, these three group fairness axioms seem to exclude a huge class of reasonable pricing
functionals. Moreover, this exclusion is insurance company-dependent, i.e., different companies may be forced
to exclude different non-protected information (due to different portfolio distributions).
This seems disputable, and it requires a broader discussion within our community
under which circumstances we should postulate one of these three group fairness axioms;
this is further highlighted in the examples of Appendix \ref{Appendix: further observations from our example}.

As a consequence, the three group fairness axioms considered here do not seem to provide a quick fix
of indirect discrimination, or they may but at a too restrictive price. In some sense, these fairness
axioms aim at mitigating discrimination in a post-hoc way which does not seem to be appropriate in every situation,
e.g., under the assumptions of Example \ref{Example}.

\medskip

We state a number of further points that are worth studying in more detail.
\begin{itemize}
\item In view of the previous conclusions, one could think of other fairness criteria that one may want to postulate.
  Alternatively, one may ask under which circumstances the group fairness axioms should be fulfilled,
and how they can complement the notion of proxy discrimination
in insurance, as the two concepts seem to be qualitatively different.
  \item Group fairness axioms are typically formulated in the binary classification case, and one
  may  explore whether the one-to-one extension to continuous responses is sensible. Generally
  speaking, the binary classification case (Bernoulli model) is a one-parameter model where the
  expected value (parameter) fully describes the distribution of the response variable. Similarly,
  there are continuous models which are fully parametrized by their expected values, and in these
  cases the group fairness criteria will essentially behave the same as in the binary classification
  case. From an actuarial perspective, Tweedie's models (belonging to the exponential dispersion family)
  are often used with the  mean function depending on covariates and a constant dispersion parameter.
  These models are examples of one-parameter models being fully described by their mean regression
  functional.

  Our example is different because the response $Y$ has an expected value of 
  $\mu(\bX)=X_1$ and a variance of $1+X_2^2$. Therefore, the response distribution is not
  fully characterized by its mean, and applying the mean functional leads to the loss of information
  of $X_2$. This may motivate a search for weaker forms of the independence, separation
  and sufficiency axioms, which do not rely on the full distribution of the response variable.  
  \item Recently, notions of conditional group fairness have gained
  more popularity. We could further explore how these behave on
  simple insurance pricing examples.
  Intuitively, these conditional versions will not essentially behave differently from our outline above,
  but they could still be useful as they study fairness criteria on smaller units of the covariate space.
  \item There are many different terms in the field of discrimination such as 
  disparate effect, disparate impact, disproportionate impact, etc.; see, e.g., Chibanda \cite{Chibanda}.  It would
  be desirable to translate these terms into more mathematical definitions.
  \item Adverse selection and unwanted economic consequences of non-discriminatory pricing should be explored.
    Non-discriminatory prices typically fail to fulfill the auto-calibration property which is crucial 
    for having homogeneous risk classes, see W\"uthrich \cite{W2022}.

\item All considerations above have been based on the assumption that we know the true model.
  Clearly, in statistical modeling there is model uncertainty which may impact different
  protected classes differently because, e.g., they are represented differently in historical
  data (statistical and historical biases). There are several examples of this type in the machine learning literature;
  see, e.g., Barocas et al.~\cite{Barocas}, Mehrabi et al.~\cite{Mehrabi} and Pessach--Shmueli \cite{PessachShmueli}.
\item Non-protected covariates may be context-sensitive. E.g., the
European Commission \cite{Europe1}, footnote 1 to Article 2.2(14) -- life
and health underwriting -- mentions the waist-to-hip ratio as a 
non-protected (useful) covariate for
health prediction. Note the following:
\begin{itemize}
\item The waist-to-hip ratio is gender dependent.
\item The waist-to-hip ratio is age dependent.
\item The waist-to-hip ratio is race dependent.
\item The waist-to-hip ratio for females depends on the number of born children.
\item Etc.
\end{itemize}
Thus, a waist-to-hip ratio $\bX$ (if non-protected) can only be correctly interpreted under the knowledge of
protected information. This will require that we pre-process non-protected characteristics
to common units. A similar situation may, e.g., occur with age, as the chronological age may be
predictive for car driving experience, whereas in life insurance we should rather choose
a biological age (which uses gender information). This raises the general issue
of how to consider non-protected information that may be context-sensitive
and may only be fully explanatory under the inclusion of protected information.
Generally, this will result in the question of how to pre-process non-protected
information.
\item We have been speaking about (non-)discrimination 
of insurance prices. These insurance prices are actuarial or statistical prices (technical premium), i.e.,
they directly result as an output from a statistical procedure. These prices are
then modified to commercial prices, e.g., administrative costs are added, 
etc. An interesting issue is raised by Thomas \cite{Thomas1, Thomas2},
namely, by converting actuarial prices into commercial prices one often distorts
these prices with elasticity considerations, i.e., insurance companies charge
higher prices to customers that are (implicitly) willing to pay more.
This happens, e.g., with new business and contract renewals
that are often priced differently though the corresponding customers may have exactly the same risk profile. In the light
of discrimination and fairness one should also clearly question such practice of elasticity
pricing as this leads to discrimination that cannot be explained by
risk profiles (no matter whether we consider protected or non-protected
information).
\item Related to the previous item on converting the technical premium into a commercial price.
  The technical premium (actuarial or statistical price) focuses on the expected response for given covariates.
  It might be that this statistical price is accepted as being free of discrimination. However,
  by converting a statistical price into a commercial price one also needs to add a risk margin (related to
  the prediction uncertainty in the statistical price), and it can well be that this risk margin (loading) adds
  discrimination to the (acceptable) statistical price.
\end{itemize}

{\small %\baselineskip.5em
\renewcommand{\baselinestretch}{.51}
}

\newpage

\appendix

\section{Appendix: further observations from our example}
\label{Appendix: further observations from our example}

We give further observations on Example \ref{Example}, but we
state them as conjectures to keep this discussion paper short (though some of them
would not be difficult to prove).

We come back to the Gaussian distribution \eqref{Gaussian model}, but we 
do not explicitly specify the values of the covariance parameters
\begin{equation*}
(X_1,X_2,D)^\top ~\sim~{\cal N} 
\left(\begin{pmatrix}0\\0\\0\end{pmatrix}, \,\Sigma=
\begin{pmatrix}
1&0&\rho_1\\
0&1&\rho_2\\
\rho_1&\rho_2&1
\end{pmatrix}
\right).
\end{equation*}
As in \eqref{Gaussian sufficiency},
for the response $Y$ we assume that conditionally, given $(\bX,\bD)=(X_1,X_2,D)$,
\begin{equation*}
Y|_{(\bX,\bD)} ~\sim~{\cal N}(X_1, 1+X_2^2).
\end{equation*}
We conjecture the following results about the three group fairness axioms.
%\begin{table}[htb]
\begin{center}
{\small
\begin{tabular}{|l||ccc|}
\hline 
 & \multicolumn{3}{|c|}{group fairness axiom}  \\
 & independence & separation & sufficiency\\
\hline\hline
$\rho_1 >0$ and $\rho_2>0$ & NO & NO & NO \\\hline
$\rho_1 >0$ and $\rho_2=0$ & NO &  NO & YES \\\hline
$\rho_1 =0$ and $\rho_2>0$ & YES & NO & NO \\\hline
$\rho_1 =0$ and $\rho_2=0$ & YES & YES & YES \\
\hline
\end{tabular}}
\end{center}
%\caption{}
%\label{fairness table}
%\end{table}
The conjectures on lines 2 and 3 are also supported by the fact that under suitable
assumptions different group fairness criteria cannot jointly hold true;
see, e.g., Chapter 3 in Barocas \cite{Barocas}.

\section{Appendix: mathematical proofs}
\label{Appendix: mathematical proofs}
We prove the above statements of Propositions \ref{proposition no independence}, \ref{proposition no separation}
and \ref{proposition no sufficiency}
in this appendix.

\bigskip

{\Beweis
  {\bf Proof of Proposition \ref{proposition no independence}.}
  Under the assumptions of Example \ref{Example} we have
  \begin{equation*}
    {\rm Cov}\left(\mu(\bX), \bD\right)=
    {\rm Cov}\left(X_1, D\right)=\rho_1=0.1>0.
  \end{equation*}
  Thus, we have a positive correlation, and \eqref{statistical parity formula} cannot hold.
This completes the proof.
  \EndProof}

\bigskip

{\Beweis
{\bf Proof of Proposition \ref{proposition no separation}.}
In view of the separation axiom \eqref{definition of separation axiom} we need
to prove, a.s.,
\begin{equation*}
\p \left[ \left. \mu(\bX) \le m \right| Y, \bD \right]=
\p \left[  \left. \mu(\bX) \le m \right| Y \right]
\qquad \text{for all $m \in \R$.}
\end{equation*}
Under the assumptions of Example \ref{Example} we have
$\mu(\bX)=X_1$ and $\bD=D$, and if the separation
axiom holds we need to have for all measurable functions $h:\R \to \R$
\begin{equation}\label{to be proved for separation 0}
\E \left[ \left. h(X_1)  \right| Y, D \right]=
\E \left[  \left. h(X_1) \right| Y \right],
\end{equation}
a.s., where these exist. We will give a counterexample to \eqref{to be proved for separation 0} which proves that the separation axiom cannot hold in
Example \ref{Example}. Because of continuity it suffices to prove that
\begin{equation}\label{to be proved for separation}
\E \left[ \left. X_1^2  \right| Y=0, D=0 \right] ~<~
\E \left[  \left. X_1^2 \right| Y=0 \right] ~<~ \E \left[X_1^2\right]=1.
\end{equation}
We give the rationale behind conjecture \eqref{to be proved for separation}. Observe that
the distribution of $X_1$ is symmetric around the origin under the Gaussian assumption \eqref{Gaussian model}.
Therefore, the right-hand side of \eqref{to be proved for separation} is precisely the variance of
$X_1$. If we condition on $Y=0$, the distribution of $X_1$ is still symmetric around the origin, therefore,
the middle term in \eqref{to be proved for separation} is the conditional variance of $X_1$, given $Y=0$.
Since $Y=0$, we expect $X_1$ (being the mean of $Y$) to be closer to zero, which implies that the conditional
variance of $X_1$, given $Y$, should be smaller than 1. Additionally conditioning on $D=0$, still keeps
$X_1$ symmetric around zero, and the positive correlation $\rho_1=0.1$ should further decrease the conditional
variance, which motivates the first inequality in \eqref{to be proved for separation}.

We start by analyzing the left-hand side of \eqref{to be proved for separation}.
We express the distribution of $(Y,X_1,X_2)$ conditionally, given $D=0$.
Using a standard result from multi-variate Gaussian distributions, see,
e.g., Corollary 4.4 of W\"uthrich--Merz \cite{WM2015}, we have conditional portfolio distribution
\begin{equation}\label{Gaussian model conditional}
\bX|_{D=0}=(X_1,X_2)^\top|_{D=0} ~\sim~{\cal N} 
\left(
\begin{pmatrix}0\\0\end{pmatrix}, \, \Sigma_D=
\begin{pmatrix}1-\rho_1^2&-\rho_1\rho_2\\-\rho_1\rho_2&1-\rho_2^2\end{pmatrix}\right).
\end{equation}
For the
response $Y$ conditionally given $(\bX,D=0)$,
we still have
\begin{equation}\label{does not change}
Y|_{(\bX,D=0)} ~\sim~{\cal N}(X_1, 1+X_2^2).
\end{equation}
Remark that \eqref{Gaussian model conditional} would coincide with the distribution of $\bX$, if we would have independence $\rho_1=\rho_2=0$,
see \eqref{Gaussian model}. Therefore, all considerations that we do under \eqref{Gaussian model conditional} can
be translated to the middle term of \eqref{to be proved for separation} by just setting these correlation parameters
equal to zero.

Thus, we aim at calculating for $h(x)=x$ and $h(x)=x^2$ the conditional expectation
\begin{equation*}
\E \left[  \left. h(X_1) \right| Y=0, D=0 \right]
=\E_{0} \left[  \left. h(X_1) \right| Y=0 \right],
\end{equation*}
where we abbreviate the conditional probability measure, given $D=0$, by $\p_0$.
This motivates us to consider
\begin{equation}\label{separation left-hand side}
\E \left[  \left. h(X_1) \right| Y, D=0 \right]
=\E_{0} \left[  \left. h(X_1) \right| Y \right]
=\E_{0}\left[\left.\E_{0} \left[  \left. h(X_1)\right| Y, X_2 \right]
\right| Y \right].
\end{equation}
The joint density of $(Y,X_1,X_2)|_{D=0} \sim f^{(0)}_{Y,X_1,X_2}$ is given by
\begin{eqnarray*}
f^{(0)}_{Y,X_1,X_2}(y,x_1,x_2)&=&
   \frac{1}{\sqrt{2\pi (1+x_2^2)}} \exp \left\{ - \frac{1}{2} \frac{(y-x_1)^2}{1+x_2^2} \right\}
   \\&& \times ~
    \frac{1}{2\pi |\Sigma_D|^{1/2}} \exp \left\{ - \frac{1}{2} 
    \frac{x_1^2(1-\rho_2^2)+x_2^2(1-\rho_1^2)+2x_1x_2\rho_1\rho_2}
    {1-\rho_1^2-\rho_2^2} \right\}.
\end{eqnarray*}    
This gives for the conditional density of $X_1$, given $(Y,X_2, D=0)$,
we drop normalizing constants in the proportionality sign $\propto$,
\begin{eqnarray*}
f^{(0)}_{X_1|(Y,X_2)}(x_1|Y,X_2)&\propto&
   \exp \left\{ - \frac{1}{2} \frac{(Y-x_1)^2}{1+X_2^2} \right\}
    \exp \left\{ - \frac{1}{2} 
    \frac{x_1^2(1-\rho_2^2)+2x_1X_2\rho_1\rho_2}
    {1-\rho_1^2-\rho_2^2} \right\}
\\&\propto&
\exp \left\{ - \frac{1}{2} \left(\frac{x_1^2-2x_1Y}{1+X_2^2} +
    \frac{x_1^2(1-\rho_2^2)
    +2x_1X_2\rho_1\rho_2}
    {1-\rho_1^2-\rho_2^2}\right) \right\}
\\&\propto&
\exp \left\{ - \frac{1}{2} \left(\frac{x_1^2
((2+X_2^2)(1-\rho_2^2)-\rho_1^2)
-2x_1\left(Y(1-\rho_1^2-\rho_2^2)
-\rho_1\rho_2X_2(1+X_2^2)\right)}
{(1+X_2^2)(1-\rho_1^2-\rho_2^2)}
\right) \right\}.           
   \end{eqnarray*}
This is a Gaussian density, and we have 
%\begin{equation*}
%    X_1|_{(Y,X_2,D=0)}~\sim~ {\cal N}\left(
% \frac{Y(1-\rho_1^2-\rho_2^2)-\rho_1\rho_2X_2(1+X_2^2)}
% {(2+X_2^2)(1-\rho_2^2)-\rho_1^2},\,
%\frac{(1+X_2^2)(1-\rho_1^2-\rho_2^2)}{(2+X_2^2)(1-\rho_2^2)-\rho_1^2} \right).
%\end{equation*}
\begin{equation}\label{Gaussian density 12}
    X_1|_{(Y=0,X_2,D=0)}~\sim~ {\cal N}\left(
 \frac{-\rho_1\rho_2X_2(1+X_2^2)}
 {(2+X_2^2)(1-\rho_2^2)-\rho_1^2},\,
\frac{(1+X_2^2)(1-\rho_1^2-\rho_2^2)}{(2+X_2^2)(1-\rho_2^2)-\rho_1^2} \right).
\end{equation}
We calculate \eqref{separation left-hand side}
for the function
$h(x)=x$, conditional on having a response $Y=0$ and $D=0$,
\begin{equation}\label{separation left-hand side h11}
\E_{0} \left[  \left. X_1 \right| Y=0 \right]
=\E_{0}\left[\left. \frac{-\rho_1\rho_2X_2(1+X_2^2)}
 {(2+X_2^2)(1-\rho_2^2)-\rho_1^2}
\right| Y=0 \right].
\end{equation}
Intuitively, this should be equal to 0 which we are going to verify.
The joint density of $(X_1,X_2)|_{Y=0, D=0}$ is given by
\begin{eqnarray*}
f^{(0)}_{X_1,X_2|Y=0}(x_1,x_2)&\propto&
   \frac{1}{\sqrt{1+x_2^2}} \exp \left\{ - \frac{1}{2}
   \left( \frac{x_1^2}{1+x_2^2} +
    \frac{x_1^2(1-\rho_2^2)+x_2^2(1-\rho_1^2)+2x_1 x_2\rho_1\rho_2}
    {1-\rho_1^2-\rho_2^2} \right)\right\}
    \\&=&
       \frac{1}{\sqrt{1+x_2^2}} \exp \left\{ - \frac{1}{2}
   \left( \frac{x_1^2((2+x_2^2)(1-\rho_2^2)-\rho_1^2)
   +2x_1 x_2\rho_1\rho_2(1+x_2^2)
   }{(1+x_2^2)(1-\rho_1^2-\rho_2^2)} +
    \frac{x_2^2(1-\rho_1^2)}
    {1-\rho_1^2-\rho_2^2} \right)\right\}.
\end{eqnarray*}    
This allows us to calculate the density of 
$X_2|_{Y=0, D=0}$ by integrating the previous density over $x_1$
\begin{eqnarray}\nonumber
f^{(0)}_{X_2|Y=0}(x_2)&=&
\int f^{(0)}_{X_1,X_2|Y=0}(x_1,x_2)\, dx_1
\\&\propto&\nonumber
\frac{1}{\sqrt{1+x_2^2}}
\frac{\sqrt{(1+x_2^2)(1-\rho_1^2-\rho_2^2)}}{\sqrt{(2+x_2^2)(1-\rho_2^2)-\rho_1^2}}
\exp \left\{ - \frac{1}{2}
     \frac{x_2^2(1-\rho_1^2)}
    {1-\rho_1^2-\rho_2^2} \right\}
\\&&\times \nonumber
\int       \frac{\sqrt{(2+x_2^2)(1-\rho_2^2)-\rho_1^2}}{\sqrt{(1+x_2^2)(1-\rho_1^2-\rho_2^2)}} \exp \left\{ - \frac{1}{2}
    \frac{x_1^2
   +2x_1 x_2\rho_1\rho_2(1+x_2^2)/((2+x_2^2)(1-\rho_2^2)-\rho_1^2)
   }{(1+x_2^2)(1-\rho_1^2-\rho_2^2)/((2+x_2^2)(1-\rho_2^2)-\rho_1^2)} \right\} dx_1
%\\&\propto&
%\frac{1}{\sqrt{(1+x_2^2)}}
%\frac{\sqrt{(1+x_2^2)(1-\rho_1^2-\rho_2^2)}}{\sqrt{(2+x_2^2)(1-\rho_2^2)-\rho_1^2}}
%\exp \left\{ - \frac{1}{2}
%     \frac{x_2^2(1-\rho_1^2)}
%    {1-\rho_1^2-\rho_2^2} \right\}
%\\&&\times 
%\exp \left\{ + \frac{1}{2}
%    \frac{ x_2^2\rho_1^2\rho_2^2(1+x_2^2)/((2+x_2^2)(1-\rho_2^2)-\rho_1^2)
%   }{(1-\rho_1^2-\rho_2^2)} \right\}
\\&\propto&\nonumber
\frac{1}{\sqrt{1+x_2^2}}
\frac{\sqrt{(1+x_2^2)(1-\rho_1^2-\rho_2^2)}}{\sqrt{(2+x_2^2)(1-\rho_2^2)-\rho_1^2}}
\\&&\times~\nonumber
\exp \left\{ - \frac{1}{2}
     \frac{x_2^2(1-\rho_1^2)((2+x_2^2)(1-\rho_2^2)-\rho_1^2)-\rho_1^2\rho_2^2x_2^2(1+x_2^2)}
    {(1-\rho_1^2-\rho_2^2)((2+x_2^2)(1-\rho_2^2)-\rho_1^2)} \right\}
\\&\propto&\nonumber
\frac{1}{\sqrt{1+x_2^2}}
\frac{\sqrt{(1+x_2^2)(1-\rho_1^2-\rho_2^2)}}{\sqrt{(2+x_2^2)(1-\rho_2^2)-\rho_1^2}}
\exp \left\{ - \frac{1}{2}
     \frac{x_2^2\left[(2+x_2^2)
     -\rho_1^2\right]}
    {(2+x_2^2)(1-\rho_2^2)-\rho_1^2} \right\}
\\&\propto&\label{we insert later}
\frac{1}{\sqrt{1+x_2^2}}
\frac{\sqrt{(1+x_2^2)(1-\rho_1^2-\rho_2^2)}}{\sqrt{(2+x_2^2)(1-\rho_2^2)-\rho_1^2}}
\exp \left\{ - \frac{1}{2}
     \frac{x_2^2(2+x_2^2)\rho_2^2}
    {(2+x_2^2)(1-\rho_2^2)-\rho_1^2} \right\}    
\exp \left\{ - \frac{1}{2}
     x_2^2\right\}.
\end{eqnarray}    
We observe that the last expression 
is symmetric w.r.t.~$x_2$ around the origin. This implies 
that the conditionally expected value in \eqref{separation left-hand side h11}
is zero. Therefore, the left-hand side of \eqref{to be proved for separation}
is equal to the conditional variance, and we have from
\eqref{Gaussian density 12}
\begin{eqnarray}\nonumber
  \E \left[ \left. X_1^2  \right| Y=0, D=0 \right]
  &=&
{\rm Var}\left(\left. X_1 \right| Y=0, D=0 \right)
~=~\E_0\left[\left.\frac{(1+X_2^2)(1-\rho_1^2-\rho_2^2)}{(2+X_2^2)(1-\rho_2^2)-\rho_1^2}
\right| Y=0 \right]
\\&=&
\frac{\E \left[\left(\frac{1}{1+X^2}\right)^{1/2}
\left(\frac{(1+X^2)(1-\rho_1^2-\rho_2^2)}{(2+X^2)(1-\rho_2^2)-\rho_1^2}
\right)^{3/2}
\exp \left\{ - \frac{1}{2}
     \frac{X^2(2+X^2)\rho_2^2}
    {(2+X^2)(1-\rho_2^2)-\rho_1^2} \right\}    
\right]}
{\E \left[\left(\frac{1}{1+X^2}\right)^{1/2}
\left(\frac{(1+X^2)(1-\rho_1^2-\rho_2^2)}{(2+X^2)(1-\rho_2^2)-\rho_1^2}
\right)^{1/2}
\exp \left\{ - \frac{1}{2}
     \frac{X^2(2+X^2)\rho_2^2}
    {(2+X^2)(1-\rho_2^2)-\rho_1^2} \right\}    
\right]},\label{separation left-hand side h3}
\end{eqnarray}
with $X \sim {\cal N}(0,1)$ and where we use \eqref{we insert later} to calculate the
corresponding integrals. That is, the denominator of \eqref{separation left-hand side h3} corresponds
to the missing normalizing constant in \eqref{we insert later}, and it can be obtained by
interpreting the corresponding integral as an expectation under a Gaussian distribution
due to the last term in \eqref{we insert later}. A similar interpretation
applies to the enumerator in \eqref{separation left-hand side h3}. From
\eqref{separation left-hand side h3} we conclude that we can calculate the
left-hand side of \eqref{to be proved for separation} by Monte Carlo simulation.

The middle term in \eqref{to be proved for separation} is calculated completely analogously,
and it can be received from \eqref{separation left-hand side h3} by setting $\rho_1=\rho_2=0$.
This results in  
\begin{equation*}\nonumber
  \E \left[ \left. X_1^2  \right| Y=0 \right]=
  {\rm Var}\left(\left. X_1 \right| Y=0 \right)
~=~
\frac{\E \left[\left(1+X^2\right)\left(\frac{1}{2+X^2}\right)^{3/2}\right]}
{\E \left[\left(\frac{1}{2+X^2}\right)^{1/2}\right]},
\end{equation*}
with $X \sim {\cal N}(0,1)$. Evaluating these terms with
Monte Carlo simulation using 10M Gaussian samples provides us with
\begin{equation*}
\E \left[ \left. X_1^2  \right| Y=0, D=0 \right]=
0.520 ~ < ~ 0.604 =  \E \left[ \left. X_1^2  \right| Y=0 \right].
\end{equation*}
This proves \eqref{to be proved for separation}, which completes the proof of
Proposition \ref{proposition no separation}.
\EndProof}

\bigskip

{\Beweis
  {\bf Proof of Proposition \ref{proposition no sufficiency}.}
Sufficiency \eqref{sufficient definition} of $\mu(\bX)$  implies that 
\begin{equation}\label{identity 0}
{\rm Var}\left(   Y  \left| \mu(\bX), \bD \right)\right.
={\rm Var}\left(  Y  \left| \mu(\bX) \right)\right..
\end{equation}
Therefore, it suffices to give a counterexample to \eqref{identity 0} under the model assumptions
of Example \ref{Example} to prove Proposition \ref{proposition no sufficiency}.
We first calculate the right-hand side of \eqref{identity 0}. It is given by
\begin{eqnarray*}
{\rm Var}\left(   Y  \left| \mu(\bX) \right)\right.
&=&{\rm Var}\left(  \left. Y  \right| X_1 \right)
\\&=&
{\rm Var}\left(  \left. \E \left[\left.Y  \right| \bX\right]\right| X_1 \right)
+\E\left.\left[{\rm Var}\left(  \left. Y  \right| \bX \right)\right| X_1 \right]
\\&=&
{\rm Var}\left(  \left. X_1\right| X_1 \right)
+\E\left.\left[ 1 + X_2^2\right| X_1 \right]
\\&=&0+1+{\rm Var}(X_2)=2,
\end{eqnarray*}
where we use that $X_1$ and $X_2$ are independent, see \eqref{Gaussian model}.
Next we calculate the left-hand side of \eqref{identity 0}. It is given by
\begin{eqnarray}\nonumber
{\rm Var}\left(   Y  \left| \mu(\bX), \bD \right)\right.
&=&{\rm Var}\left(  \left. Y  \right| X_1, \bD \right)
\\&=&\nonumber
{\rm Var}\left(  \left. \E \left[\left.Y  \right| \bX, \bD\right]\right| X_1, \bD \right)
+\E\left.\left[{\rm Var}\left(  \left. Y  \right| \bX, \bD \right)\right| X_1, \bD \right]
\\&=&\nonumber
{\rm Var}\left(  \left. X_1\right| X_1, \bD \right)
+\E\left.\left[ 1 + X_2^2\right| X_1, \bD \right]
\\&=&0+1+\E\left.\left[ X_2^2\right| X_1, D \right].\label{last term 1}
\end{eqnarray}
The distribution of $X_2$, given $(X_1,D)$, can be calculated under the Gaussian assumption \eqref{Gaussian model},
and it is given by
\begin{equation*}%\label{posterior needs some calculation}
  X_2|_{(X_1,D)}~\sim~ {\cal N}\left(
    \frac{\rho_2}{1-\rho_1^2}\left(D-\rho_1X_1\right),\,\frac{1-\rho_1^2-\rho_2^2}{1-\rho_1^2}
    \right),
  \end{equation*}
this can be obtained, e.g., from Corollary 4.4 of W\"uthrich--Merz \cite{WM2015}.
This implies that in general the last term in \eqref{last term 1} is different from 1 because it
depends on $(X_1,D)$.
  Henceforth, identity \eqref{identity 0} cannot hold, 
which completes the proof.
  \EndProof}

\end{document}